\documentclass[conference]{IEEEtran}
\IEEEoverridecommandlockouts
\usepackage[square,numbers]{natbib}
\usepackage{amsmath,amssymb,amsfonts}
\usepackage{algorithmic}
\usepackage{graphicx}
\usepackage{textcomp}
\usepackage{xcolor}
\usepackage{subfig}
\def\BibTeX{{\rm B\kern-.05em{\sc i\kern-.025em b}\kern-.08em
    T\kern-.1667em\lower.7ex\hbox{E}\kern-.125emX}}

\begin{document}

\title{Human-Machine Teaming For UAVs: An Experimentation Platform}

\author{\IEEEauthorblockN{Laila El Moujtahid}
\IEEEauthorblockA{\textit{AI Redefined} \\
Montreal, Canada \\
laila@ai-r.com}
\and
\IEEEauthorblockN{Sai Krishna Gottipati}
\IEEEauthorblockA{\textit{AI Redefined} \\
Montreal, Canada \\
sai@ai-r.com}
\and
\IEEEauthorblockN{Clodéric Mars}
\IEEEauthorblockA{\textit{AI Redefined} \\
Montreal, Canada \\
cloderic@ai-r.com}
\and
\IEEEauthorblockN{Matthew E. Taylor}
\IEEEauthorblockA{\textit{AI Redefined} \\
Montreal, Canada \\
matt@ai-r.com}
}

\maketitle

\begin{abstract}
Full automation is often not achievable or desirable in critical systems with high-stakes decisions. Instead, human-AI teams can achieve better results. To research, develop, evaluate, and validate algorithms suited for such teaming, lightweight experimentation platforms that enable interactions between humans and multiple AI agents are necessary. However, there are limited examples of such platforms for defense environments. To address this gap, we present the Cogment human-machine teaming experimentation platform, which implements human-machine teaming (HMT) use cases that features heterogeneous multi-agent systems and can involve learning AI agents, static AI agents, and humans. It is built on the Cogment platform and has been used for academic research, including work presented at the ALA workshop at AAMAS this year~\cite{thunderblade_aamas_ala}. With this platform, we hope to facilitate further research on human-machine teaming in critical systems and defense environments. 
\end{abstract}


\section{Introduction}


\begin{figure*}[h]
\centerline{\includegraphics[clip, width=0.9\textwidth]{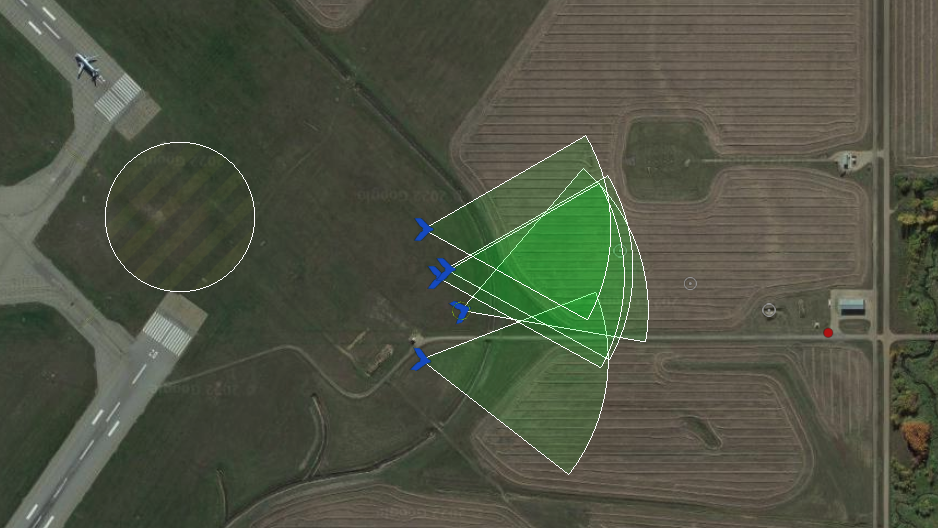}}
\caption{\label{fig:user_interface}This figure shows the main user interface for the Cogment HMT experimentation platform. The hatched circle on the left is a restricted area that the team of five \textit{blue} agent defends. On the right side, the single red dot is the UAV attacker.}
\end{figure*}

Embodied AI agents, such as unmanned aerial vehicles (UAVs, or drones), have the potential to revolutionize various industries, including transportation, agriculture, and security. However, these agents evolve in the physical world and, as such, can have dangerous effects, especially when left unsupervised. For instance, a UAV may malfunction or fail to identify potential hazards, resulting in property damage or even human injury. Moreover, embodied AI agents can make decisions based on algorithms that may not consider ethical, moral, or legal implications. Therefore, it is essential for humans to have the ability to exercise meaningful control~\cite{Meaningful_human_control_1} and oversight on these agents to ensure their safe and responsible use. Human operators can monitor and intervene in cases of system malfunction, assess potential risks, and make ethical or legal decisions in complex situations that require their judgment. 

In addition to oversight, humans can play a critical role in helping embodied AI agents achieve their tasks through collaboration. For example, in the case of UAVs, human operators in control centers can provide real-time guidance and support, ensuring that they perform the desired functions accurately and efficiently. Furthermore, humans can act as teammates in the field, working alongside embodied AI agents to achieve complex goals that require both human judgment and machine precision.

Moreover, it is important to recognize that human-machine teaming (HMT), the ability for humans and embodied AI agents to create a bidirectional collaboration, is a key aspect of safe and effective use of AI. The design, training, validation, and operation of such AI agents cannot be done in isolation; it is important to consider how they fit into a larger system that includes them. Humans, in particular, as operators or as teammates, should be considered an integral part of this system from the beginning.

Beyond this bidirectional collaboration, embodied AI systems often fail to consider ``moral responsibility'' and ``socio-technical'' factors in their operation~\cite{Meaningful_human_control_1}. The concept of meaningful human control (MHC) was introduced by \citeauthor{Meaningful_0} to enable humans to influence the behavior of embodied AI agents~\cite{Meaningful_0}. However, the original definition of MHC is inconsistent because humans may lack the expertise or the knowledge to fully control AI systems effectively. \citeauthor{Meaningful_human_control_1}~\cite{Meaningful_human_control_1} proposed four additional properties to improve the original definition of MHC: an ``explicit moral operational design domain,'' ``appropriate and mutually compatible representations,'' ``control ability and authority,'' and ``explicit linkage between AI and human actions.''
Therefore, it is crucial to design an orchestration platform to combine meaningful human control and human-in-the-loop to ensure that AI systems are trained and operated in a way that aligns with human values, societal norms and ethical behavior.

Cogment HMT provides a platform to design and experiment with human-machine teams, in particular involving UAVs. Built upon our Cogment~\cite{cogment} platform, it addresses the challenges of orchestrating collaboration between automated decision-making systems, including AI agents, humans, and their access to data and their effects on their environment. The Cogment HMT experimentation platform currently uses one simulated environment, and can be easily adapted for more realistic simulations and real-world deployments. We describe it and its properties in Section \ref{sec:platform}. 

Using the Cogment HMT experimentation platform, AI practitioners can develop agents capable of collaboratively working with humans and learn from their knowledge and expectations, and consider factors such as meaningful human control, trust, and managing cognitive load and enable effective bidirectional human-machine collaboration. We describe early results in Section \ref{sec:results}. 

\section{Related work}

\subsection{Autonomous robotics}

Autonomous robotics has seen remarkable progress in recent years with the emergence of various popular experimental environments, including Gym~\cite{Gym}, MuJoCo~\cite{mujoco}, and IsaacGym~\cite{IsaacGym}. These environments are essential tools for researchers and developers to design and test algorithms quickly and easily in complex robotic ecosystems. Gym~\cite{Gym} is an open-source toolkit, developed by Open AI, and widely used in academia and industries. It provides a standardized interface and tools for creating, benchmarking, evaluating reinforcement learning algorithms. on a wide range of tasks, such as playing Atari games or controlling robots. Other relevant robotics simulation environments, such as Robosuite~\cite{robosuite} built on top of MuJoCo, and IsaacGym, were developed to cover the need for physics-based simulation in robotics and computation requirements. To accurately model and simulate the physical interactions between robotic systems and their real-world environments, it is crucial for researchers and developers to model and capture the underlying physical laws and principles governing the behavior of objects in the real world such as gravity, friction, and collisions.

These experimentation platforms are complementary and well-suited for tasks that require a high degree of physical realism and significant computational resources, particularly for simulating large-scale robotic ecosystems. By using these benchmarks, researchers and developers can more easily identify and compare the strengths and weaknesses of different algorithms and select the most effective approaches in a controlled, safe, and cost-effective manner. However, these platforms focus on single robots, and are not suited for collaboration at scale.

\subsection{Multi-agent Systems}

 Multi-agent reinforcement learning (MARL)~\cite{RL} is a sub-field of reinforcement learning (RL)~\cite{Sutton1998} that focuses on learning algorithms for multi-agent systems. MARL is a relatively new area of research in artificial intelligence and robotics. In contrast to a single agent RL, MARL considers systems in which multiple agents interact with each other and the environment to achieve a common or individual goal~\cite{wong2022deep}. Each agent operates independently and makes decisions based on its local observations, which may affect the observations and decisions of other agents~\cite{zhang2021multi}. Relatively few experimentation platforms support these types of systems, making it challenging for researchers and developers to test and evaluate their algorithms. PettingZoo~\cite{pettingzoo} is an open-source experimentation platform. It supports heterogeneous and homogeneous multi-agent systems. It provides a collection of benchmark environments that simulate real-world scenarios, with a well-designed API and libraries, and agent interface tooling. However, when it comes to human teaming, these experimental environments have some limitations. Libraries, tools, and interfaces are not suitable for allowing agents to interact and collaborate with humans.

 Beyond the platforms, multi-agent reinforcement learning (MARL) poses considerable challenges when it is time to design a robust and intelligent system ~\cite{MRRL_Challenges}: 
 \begin{enumerate}
    \item Scalability: The number of agents in a system can increase the complexity of learning and coordination between the agents, making it challenging to develop scalable algorithms for MARL.
    \item Non-stationarity: The behavior of other agents in dynamic environments can change over time, making it difficult to develop learning algorithms that can adapt to these changes. 
    \item Communication and coordination: Designing adequate protocols and coordination mechanisms that can handle the heterogeneity of agents~\cite{Multi_Robot_RL_heterogeneous}. This becomes complicated when introducing robot-human social interaction and hardware requirements. The algorithms approaches need to be designed to handle the heterogeneous nature of the agents, as well as the dynamic and uncertain nature of the environments they operate in.
    \item Partial Observation: An agent may not have access to complete information about the environment. The trade-off balance between exploration and exploitation can be delicate for each agent strategy while also exploiting their knowledge to achieve isolated tasks individually in collaborative interaction.
    \item Reward function: Designing a reward function that encourages agents to work together while considering their heterogeneity and capabilities can also be a challenging task.
 \end{enumerate}
 
Finally, existing MARL platforms are designed to support relatively simple and small-scale environments, where agents have to learn simple tasks in homogeneous systems~\cite{Multi_Robot_RL_heterogeneous}. The success of MARL algorithms in one environment does not guarantee its success in another environment, making it challenging to develop algorithms or platforms that can bidirectionally transfer information or knowledge to different environments. Therefore, when it comes to designing more complex or realistic heterogeneous multi-robot or multi-agent systems, a more advanced platform with powerful computation and algorithmic capabilities is needed.

\subsection{Human-Machine Teaming}

Human-machine teaming (HMT) has become increasingly important in many fields, including robotics, aviation, and defense. This approach refers to the bidirectional collaboration between humans and multi-agent systems through a cognitive interface. There are several different approaches to training agents for HMT including the following human-in-the-loop learning (HILL) techniques: 
\begin{enumerate}
    \item Training from human demonstration refers to a method of teaching a machine how to perform a task by having a human demonstrate it. The machine can then observe and learn from the human's actions, and use this information to perform the task autonomously~\cite{pomerleau1988alvinn}.
    \item Training from human feedback, or reinforcement learning from human feedback (RLHF), refers to a process where humans provide guidance and correction to machines in order to evaluate the behavior of the system and provide feedback to help the machine improve its performance~\cite{christiano2017deep}.
    \item Human intervention refers to the practice of a human operator taking control of a machine in order to modify its behavior, either by overriding its actions or by enriching the learning with new tasks~\cite{Chernova2009}. 
\end{enumerate}
A comprehensive and detailed exploration of how humans can assist Reinforcement Learning agents in their learning process can be found in~\cite{Conceptual_Framework}.

These approaches are applied in various applications, such as the Hanabi card game~\cite{Hanabi}. This game is used as a research tool to study collaboration and communication between humans and machines, as it requires players to work together to achieve a common goal. Originally designed as a benchmark for training cooperative agents, Hanabi has also been used to measure zero-shot training capabilities in collaborating with new players~\cite{Hanabi-Ref12}. Ongoing research, such as the work done in Cogment-Verse~\cite{cogment_verse}, involves implementing the entire pipeline of self-play, training with randomly selected pre-trained agents, and testing and improving zero-shot coordination capabilities by involving human players during training or testing phases. Cogment's ability to efficiently switch between different actors, run multiple parallel trials, and support different training paradigms simultaneously has proven helpful in this endeavor. Another relevant example of a human-in-the-loop implementation is collaborative robots, or cobots~\cite{COBOT_MIT}. These robots are designed to work alongside human operators in a collaborative environment, sharing tasks and responsibilities to improve efficiency, productivity, and safety. For example, cobots can perform repetitive tasks that may be difficult or unsafe for humans to perform on their own~\cite{Exploring_Human-robot_Interaction_Simulating_Robots}. In other cases, cobots can assist human operators by handing them tools or materials, or by holding parts in place while the human performs assembly or other tasks. To achieve these results, these robots are equipped with sensors and other technologies that allow them to perceive the environment and interact with humans in a natural and intuitive way.

\subsection{Multi-agent UAV Platforms}

Only a few frameworks aim to facilitate research and development of UAV teams. Aerostack~\cite{aerostack} is an open source software platform based on ROS~\cite{ros2} that enables developers to control and test UAVs in a specific context of inspection and surveillance. It interfaces with a variety of APIs and tools, incorporating a library of computer vision algorithms and sensors. It provides a modular, and customized multi layout architecture including modules level such as behavior, embedded controls and functional module interfacing through Aerostack APIs and services. While Aerostack provides strong equipment for human-robot interaction, there are some limitations when it comes to incorporate human feedback through HILL approaches. The platform provides useful tools and services, but it is unable to incorporate human feedback or human demonstrations. We believe such mechanisms are critical when transitioning from simulation to real-world applications. 

Gazebo~\cite{gazebo} is another 3D robot simulation software used for modeling and simulating robots, sensors, and environments. It is also an open-source software that is widely used in robotics research, and industrial robots. Gazebo provides a comprehensive set of tools and libraries for simulating UAVs, including flight controllers, sensors, and communication interfaces, and can be integrated with ROS for additional functionality. While Gazebo is a useful tool for simulating physic-robotic environments, it has its limitations when it comes to model complex UAV behaviors and social interactions within a multi-environment, such as interaction with operators and others UAVs in the real world, with a high degree of accuracy.

\section{Platform}
\label{sec:platform}

Cogment~\cite{cogment} is a general solution designed to build, train, and operate AI agents in simulated or real environments shared with humans. We used Cogment to implement an experimentation platform enabling AI research \& development in scenario involving the collaboration between humans and AI-driven UAVs in defense and security use cases. Furthermore, the Cogment HMT experimentation platform provides a smooth path towards deployment in the real world.

\subsection{Simulation}

The Cogment HMT experimentation platform is designed to experiment with scenarios involving a teams of UAVs collaborating with humans. In its current form, the platform is focused on defensive scenarios to safeguard critical infrastructures. UAVs are operated from a ground control station by a human operator who focuses on high-level tasks such as threat detection and interception. Low-level control of the flight dynamics is not within the scope of this work --- we currently assume the speed of computers will outperform a human's direct operation of a UAV.

\begin{figure}[htbp]
\centerline{\includegraphics[width=0.95\columnwidth]{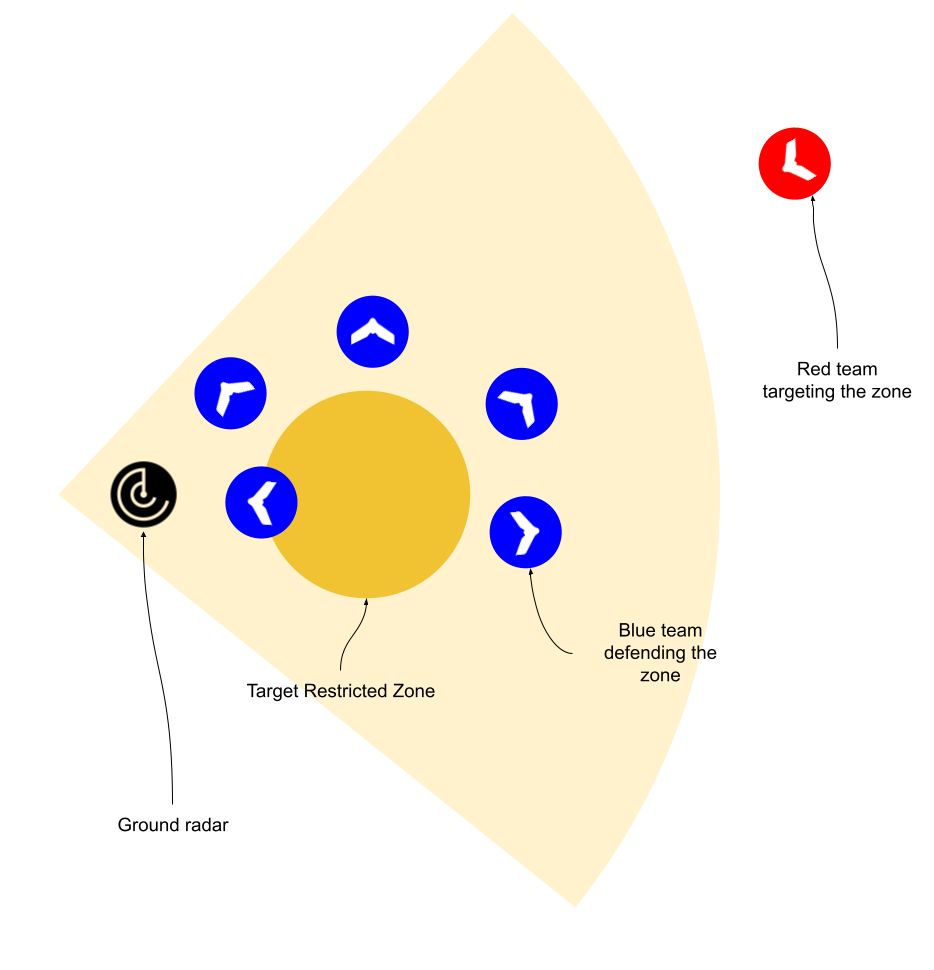}}
\caption{This schematic representation of a typical scenario in the Cogment HMT experimentation platform showcases a \textit{blue} team composed of 5 UAVs defending a restricted zone from intrusion by a \textit{red} team of 1 UAV.}
\label{fig:scenario}
\end{figure}

The platform supports scenarios where two teams are present: the \textit{red} team aims to penetrate a restricted zone, while the \textit{blue} team aims at protecting this zone by intercepting the intruders. Figure~\ref{fig:scenario} presents a schematic representation for these types of scenario.

Each team consists of one or multiple UAVs. Simplified flight dynamics are implemented for fixed-wing, vertical take-off and landing. Because flight control is not the focus of the platform, the modeling is simplified and projected into two dimensions. Each drone can be equipped with multiple sensors defined by their characteristics including range, orientation, and probability to detect. Finally, the UAVs can have a payload that can vary in their level of dangerousness --- in our scenario, the red team has no payload and the blue team have electromagnetic pulse devices.

Additional fixed (e.g., ground-based) sensors can be added to the scenario to help the \textit{blue} team with threat detection. They can have the same properties than the UAV sensors.

Each entity, UAV or fixed sensor, can be set up to have a full awareness of the other entities, automatically share information with their teammates, or only rely on their sensors. This allows researchers to consider offensive and defensive tasks as fully or partially observable, with or without communication. Entities' dynamics, sensory capabilities, and payloads are simulated in a dynamically configurable simulation that is integrated as an environment in a Cogment application. 

\subsection{Multi-agent architecture}

\begin{figure*}[htbp]
\centerline{\includegraphics[width=0.8\textwidth]{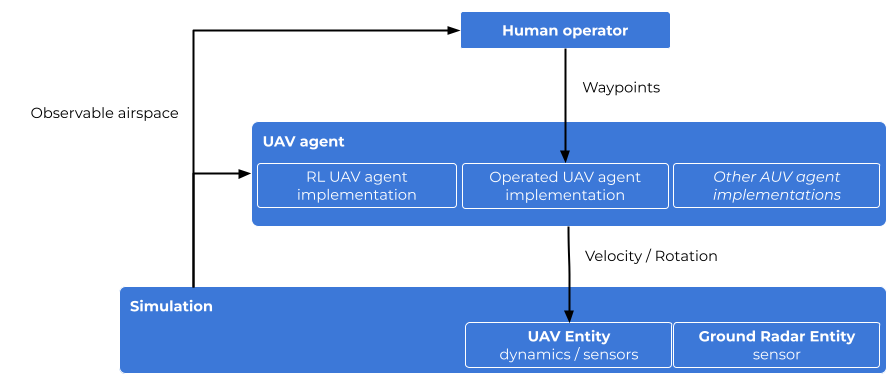}}
\caption{Multi-agent architecture}
\label{fig:multi_agent_architecture}
\end{figure*}

The platform models the scenarios as a multi-agent system, described in Figure~\ref{fig:multi_agent_architecture}. The primary agent type controls the UAVs through velocity and rotation changes. Agents receive full or partial observations from the environment. Multiple implementations of UAV agents can be defined and instantiated, including path following, heuristic behaviors, or trained policies. Furthermore, the agent can dynamically switch between multiple control strategies over time.

The other type of actor enables human operators to have indirect control over individual UAVs. This allows a dynamic human-AI team-up in training or operation phases. The operator actor can add and remove waypoints to any UAV agent in their team. The inputs are then provided to the UAVs for decision making. For example, the operated UAV agent implementation follows a simple path following policy which consumes waypoints. 

\subsection{Platform use cases}
The experimentation platform implementation choices are based on its use cases. To conduct the experiments described in Section~\ref{sec:results}, four use cases have been identified:

\begin{enumerate}[]
\item Record demonstrations from human operators who fully control the drones.
\item Run MARL non-interactive batch training leveraging pre-recorded demonstrations.
\item Operate scenarios involving both human operators and trained AI agents for test and validation purposes.
\item Run MARL online involving human operators.
\end{enumerate}

The experimentation platform runs episodes (i.e., instances of the airspace simulation), where one team is the winner at the end of every episode. Use Cases 1, 3, and 4 require running interactive episodes when human operators connect to the experimentation platform. Use case 2 requires the ability to run a large number of non-interactive (headless) episodes driven by the agents' training process. Use case 4 requires the ability to run batches of interactive episodes, and potentially headless episodes, and use the resulting data feed to continually train the agents.

\subsection{Cogment-based implementation}

\begin{figure*}[htbp]
\centerline{\includegraphics[width=0.8\textwidth]{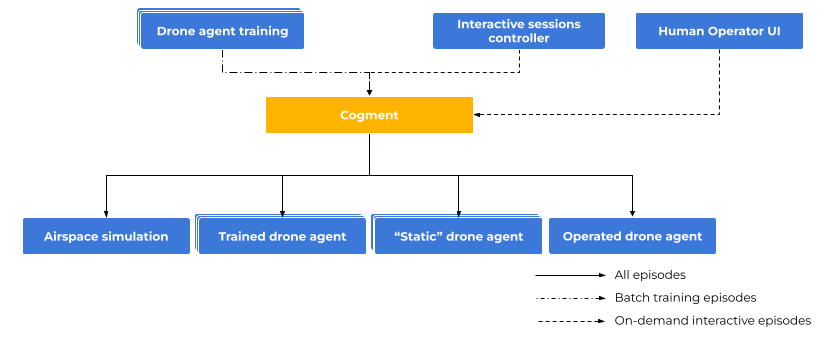}}
\caption{Components architecture}
\label{fig:components_architecture}
\end{figure*}

From the identified use cases we designed and implemented the Cogment HMT experimentation platform architecture represented in Figure~\ref{fig:components_architecture}. Cogment is used to handle the orchestration of the execution and communication between the different components:
\begin{itemize}
    \item The agents, encapsulated in dedicated µServices as Cogment actors
    \item The simulation, encapsulated in a dedicated µService as a Cogment environment
    \item The human operator UI, encapsulated as a client Cogment actor
    \item The training process, as a python script relying on the Cogment SDK to trigger trials and retrieve generated activity data 
    \item The interactive session controller, as a component of the frontend, relying on the Cogment SDK to trigger episodes
\end{itemize}

Cogment dispatches observations of the environment from the simulation to the agents, as well as instructions from higher-level agents. It then dispatches agents’ actions to the environment, which updates the simulation and agents' instructions. Furthermore, a priority-ordered list of multiple agents can be assigned to a single drone entity at once. If a higher priority agent outputs a velocity and rotation change, it overrides lower priority ones (e.g., enabling dynamic takeover by the human operator).

Cogment also provides its trial datastore, storing and making available the activity data generated by all actors in all environments.

The platform has been successfully deployed natively on workstations on both Linux, MacOS, and Windows. Longer headless training were executed on Compute Canada's slurm-based system. Finally, it was deployed on cloud infrastructure on AWS Canadian datacenter to experiment with human input.

\subsection{Interactive Interface}

The frontend part of the experimentation platform is a web application built in Javascript using React as its main framework. It consists of two different screens: an episode configuration form and an interactive episode runtime view. 

The episode configuration screen enables users to configure different parameters. The episode runtime view in Figure~\ref{fig:user_interface} shows a map of the environment with a top view, the drones and their detection range, the ground radar, and  detected red drones. Each drone can be selected by clicking on it; from there, users can add a waypoint for this drone by clicking on the map or remove existing waypoints. Finally, users can pan and zoom the map and pause or resume the simulation. This simple, yet polished user interface enables a high degree of interactivity and can be easily extended to support other tasks or collaboration modalities.

\section{Experiments}
\label{sec:results}

The experimentation platform was initially developed to support recent research~\cite{thunderblade_aamas_ala}. In this work, we investigate the following two research questions:
\begin{enumerate}[]
    \item How well does a trained agent perform in this specific environment? 
    \item Do agent or human demonstrations help to make the RL agent more sample efficient? 
\end{enumerate}

In this section, we summarize this work to illustrate how the platform is leveraged. 

We aim to train RL policies for the blue drone with the following Markov decision process formalization. The observation space consists of the relative positions of the drones and restricted airspace over three time steps. The action space is discrete and the reward function is positive for blue drones successfully neutralizing the red drone and negative if the red drone enters the restricted zone. The blue drones also receive a shaping reward proportional to their relative distance from the red drone in consecutive time steps. 

The agents are trained using multi-agent centralized training and execution setup, with each agent having its own observation space and the same reward function and action space. 

We evaluated the performance of the trained agent using the \textit{success rate} as the metric. The success rate is defined as the percentage of times the blue team wins over all evaluation trials. This metric was chosen as it provides a clear and intuitive measure of the agents' ability to defeat the red drone and is directly proportional to the average reward. We run thirty evaluation episodes per hundred training episodes to compute the success rate. During the evaluation episodes, agents do not learn or explore. Our learning curves show the performance metric reported in the evaluation episodes averaged over five runs with different seed values to account for training and environmental stochasticity.

\begin{figure}[h]
  \centering
  \includegraphics[width=\columnwidth]{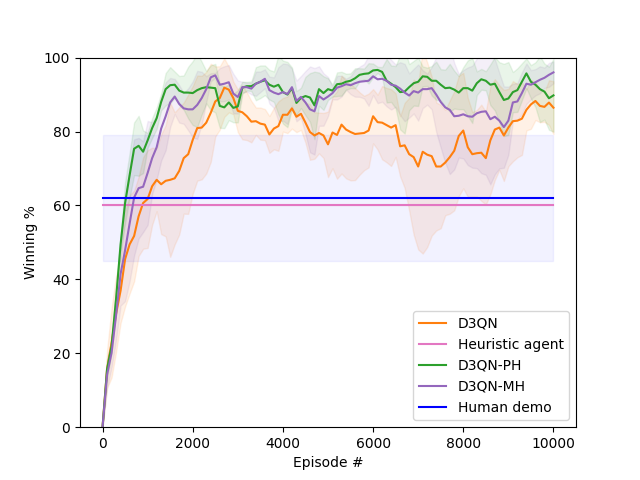}
  \caption{The success rate comparison for D3QN, D3QN with trained agent demonstrations, D3QN with real human and trained agent mix demonstrations, and a heuristic baseline. Here, the suffix -PH represents demonstrations from a trained agent and -MH indicates a mixture of real and trained agent demonstrations. 
  } 
  \label{fig:performance}
\end{figure}

\begin{figure*}[h]
   \subfloat[trained agent demonstration]{
      \includegraphics[clip, width=0.3\textwidth]{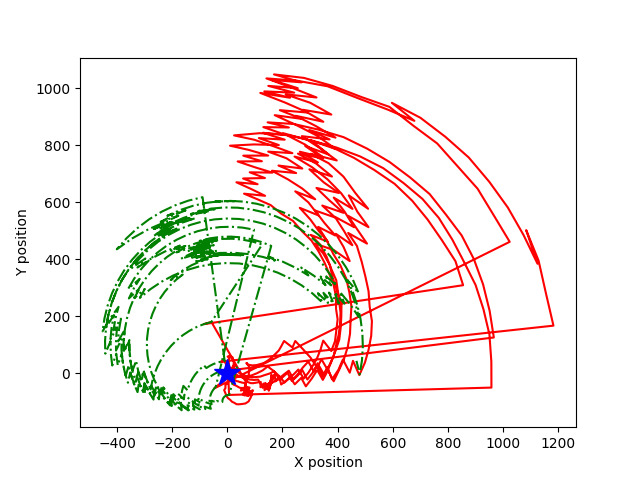}
      }
      \label{human_demo}
\hspace{\fill}
   \subfloat[a human user (User~$1$)\label{pyramidprocess} ]{%
      \includegraphics[clip, width=0.3\textwidth]{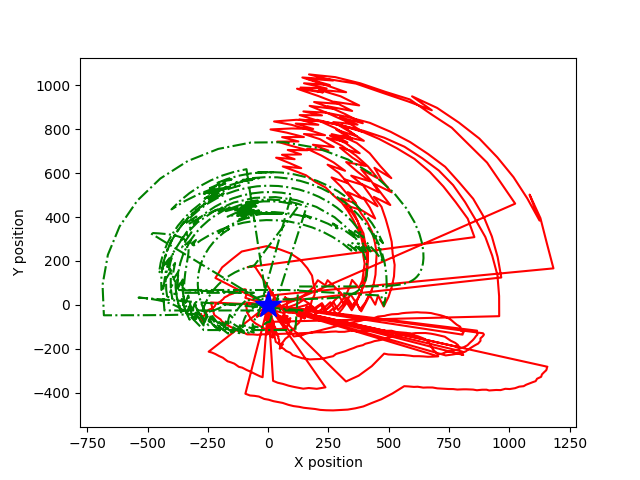}}
\hspace{\fill}
   \subfloat[a human user (User~$2$)\label{mt-simtask}]{%
      \includegraphics[clip, width=0.3\textwidth]{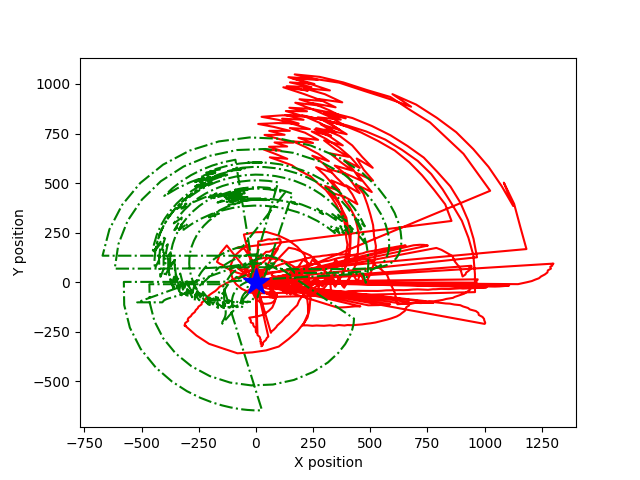}}\\
\caption{\label{fig:human_demo}Visual representation of five episodes from trained agent and two different real human users}
\end{figure*}

We train a D3QN agent (cite) and plot its performance in Figure~\ref{fig:performance}. The agent reaches a success rate of roughly $90\%$ in $3500$ episodes. The trained agent outperforms the baseline heuristic agent, which has a success rate of $60\%$. 
The average human performance is around $63\%$, which is almost equal to the heuristic agent. 

D3QN-PH reaches a success rate of more than $90\%$ in $1600$ episodes, outpacing D3QN, as shown in Figure~\ref{fig:performance}. We performed an unpaired t-test to examine the significant difference between the performance of the D3QN-PH and D3QN agents. The D3QN-PH agents exhibit a statistically significant performance improvement
compared to D3QN ($p < 0.0001$) and the effect size is $1.43$. At the end of learning, both algorithms converge to the same final performance level (around $90\%$). 
This supports our claim that trained agent demonstrations make the RL agent more sample efficient in our environment, consistent with existing results in the literature~\cite{hester2018deep,nair2018overcoming}.

\subsection{Pilot study using human demonstration}

We also trained the learning agent with a mix of agent and actual human demonstrations: D3QN-MH in Figure~\ref{fig:performance}. We sampled an equal proportion of human and trained agent demonstrations in every mini-batch used for training. We used a mix of both types of demonstrations due to the lack of human demonstrations collected in our ongoing pilot study. The results do not suggest a significant learning improvement compared to D3QN-PH; however, the performance is still statistically significantly improved over the D3QN  baseline.

We visualized five trajectories of all blue and red drones from trained agent demonstrations and two actual human demonstrations from different users who played more than $30$ games, as shown in Figure~\ref{fig:human_demo}. In this figure, the blue star denotes one of the ally drones that neutralized the enemy drone and is the frame of reference (located at $(0,0)$). The red and green lines represent the relative position of the enemy drone and the restricted airspace (with respect to the blue drone's position). Figure~\ref{fig:human_demo}(a) shows five trajectories of ally drones generated from the trained D3QN agent. 
We note that across all the figures, the red drone starts moving towards the restricted zone
while being chased by the blue drones until it is neutralized. The low density of red lines around the blue star indicates that the blue drones quickly neutralizes the red drone without following it for a long time.  
From Figure~\ref{fig:human_demo}(b) and ~\ref{fig:human_demo}(c), which depicts trials by two different human participants, we notice that there is more movement (high-density) around the blue star, suggesting that the human tries setting waypoints in different areas (using the whole team of five blue drones) of the map to neutralize the red drone. These trajectories are sub-optimal (longer trajectory length) as compared to the trajectories from trained agent demonstrations. However, these might be helpful to neutralize the red drones in difficult environment configurations where the trained RL agents fails to catch the enemy drone (trained agents have a failure rate of around $10\%$ in this task). 

\section{Future work}

Since this simulation has a learning curve for humans because of the interface and the dynamics, we intend to collect more demonstrations from humans and to include a burn-in period for humans to understand the environment and learn to play before collecting demonstrations.

In order to advance research and development on human-machine teaming, we aim to continue expanding the experimentation platform.   

We are working on accelerating the time to experimentation by providing access to off-the-shelf algorithms implementing well-known HILL learning techniques. This will allow researchers to focus on the core challenges of their research, rather than on implementing the basics of the learning loop.

We want to encompass a wider range of scenario types. This will enable researchers and developers to explore the potential of human-machine collaboration in more domains in defense and beyond, such as disaster response or logistics. One example of extension is the ability to support hybrid teams in the air including both UAVs and manned aircrafts.

Developing collaborative systems involving embodied AIs and human operators starts in simulation and aims at being deployed in the real world: the \textit{sim-to-real} challenge. While the experimentation platform primarily aims at supporting \textit{in silico} development and experimentation, by leveraging Cogment it is ready to move towards real world deployment. 

\textit{Sim-to-real} is not a one-step process from an agent developed in simulation to the real world. In fact, creating intermediate steps can make this process smoother and less challenging. Because it decouples the actor and environments' roles with their instances, Cogment makes this process smoother. The Cogment HTM experimentation platform leverages this feature to support both fully simulated episodes and interactive episodes, going from simulated \textit{pseudo-human} actors to actual human actors. At the same time, it can go from its current simple airspace simulation to more realistic one and finally to pilot actual UAVs. 

To ensure that the results obtained from our experimentation platform are reliable and can be compared across different studies, we aim at developing standardized evaluation methods. This is a challenging task in itself, as meaningful evaluation of human-machine teaming requires a holistic understanding of the role of humans in the loop. Evaluation metrics need to take into account not only task performance, but also aspects such as human trust, situation awareness, workload, and cognitive load.

Finally, we recognize that the interface between humans and AI agents is a crucial aspect of human-machine teaming. We are exploring how natural language interfaces can facilitate communication and collaboration between humans and AI agents, with the goal of improving trust, transparency, and robustness. This is a challenging research area, as natural language interfaces for complex systems require sophisticated natural language processing, dialogue management, and reasoning capabilities. However, we believe that this research can pave the way towards more effective and intuitive human-machine teaming interfaces in the future.

\section{Conclusion}

The continued exploration and research around human-machine teaming systems has the potential to revolutionize a wide range of applications, from robotics to aerospace and defense. Emerging technologies such as large language models, MARL, and embodied robotics are playing an increasingly significant role in reinventing the \textbf{bidirectional collaboration} between humans and machines for decision-making in complex and critical environments~\cite{thunderblade_aamas_ala}.

In this context, the Cogment HMT experimentation platform provides a powerful and flexible tool for applied research and evaluation of HMT systems. Early results leveraged the main benefits of the platform including a simple and fast simulation, heterogeneous multi-agent architecture, the ability to run both headless and interactive episodes, and the dynamic agent ``takeover'' capability. The platform's flexibility enables researchers to easily test different configurations and make adjustments without affecting the overall functioning of the system. 

We hope that this work and Cogment can facilitate the acceleration of human-machine teaming development while keeping the human factors at the center of this effort.

\section*{Acknowledgment}

The authors would like to express their gratitude to the members of the Intelligent Robot Learning lab at the University of Alberta, JACOBB.AI, and Thales for their collaboration on this project. The authors would also like to acknowledge the financial support of the Natural Sciences and Engineering Research Council of Canada and the PROMPT program of the Quebec province.

\bibliographystyle{IEEEtranN} 


\end{document}